\documentclass[10pt,twocolumn,letterpaper]{article}

\usepackage{iccv}
\usepackage{times}
\usepackage{epsfig}
\usepackage{graphicx}
\usepackage{amsmath}
\usepackage{amssymb}
\usepackage{amsfonts}

\usepackage{booktabs}
\usepackage{tabularx}
\usepackage{enumitem}
\usepackage{adjustbox}
\setlist{nosep}
\usepackage[normalem]{ulem}

\newcolumntype{Y}{>{\raggedleft\arraybackslash}X}
\newcolumntype{L}{>{\hsize=1.25\hsize}X}
\newcolumntype{s}{>{\hsize=.5\hsize}X}

\newcommand{\Tcal}{{\mathcal{T}}}

\newcommand{\myparagraph}[1]{\vspace{2pt}\noindent\textbf{#1.}}

\usepackage[pagebackref=true,breaklinks=true,letterpaper=true,colorlinks,bookmarks=false]{hyperref}

 \iccvfinalcopy 

\def\iccvPaperID{3496} 

\newcommand{\jp}{JPoSE}

\newcommand{\ip}{PoS-MMEN}

\newcommand{\mm}{MMEN}

\ificcvfinal\fi
\begin{document}

\title{Fine-Grained Action Retrieval Through Multiple Parts-of-Speech Embeddings}

\author{Michael Wray\\University of Bristol
\and Diane Larlus\\Naver Labs Europe
\and Gabriela Csurka\\Naver Labs Europe
\and Dima Damen\\University of Bristol}

\maketitle
\thispagestyle{empty}

\begin{abstract}
  We address the problem of cross-modal fine-grained action retrieval between text and video.
  Cross-modal retrieval is commonly achieved through learning a shared embedding space, that can indifferently embed modalities.
  In this paper, we propose to enrich the embedding by disentangling parts-of-speech (PoS) in the accompanying captions.
  We build a separate multi-modal embedding space for each PoS tag. The outputs of multiple PoS embeddings are then used as input to an integrated multi-modal space, where we perform action retrieval.
  All embeddings are trained jointly through a combination of PoS-aware and PoS-agnostic losses.
  Our proposal enables learning specialised embedding spaces that offer multiple views of the same embedded entities.
  
  We report the first retrieval results on fine-grained actions for the large-scale EPIC dataset, in a generalised zero-shot setting. Results show the advantage of our approach for both video-to-text and text-to-video action retrieval. 
We also demonstrate the benefit of disentangling the PoS for the generic task of cross-modal video retrieval on the MSR-VTT dataset.
\end{abstract}


\section{Introduction}

\begin{figure}
    \centering
    \includegraphics[width=\linewidth]{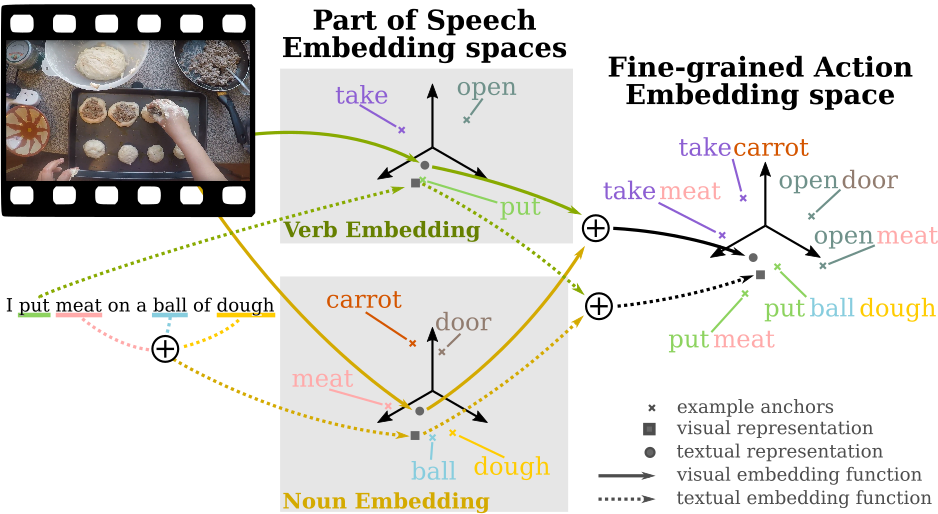}
    \caption{We target fine-grained action retrieval. Action captions are broken using part-of-speech (PoS) parsing. We create separate embedding spaces for the relevant PoS (\eg Noun or Verb) and then combine these embeddings into a shared embedding space for action retrieval (best viewed in colour).} 
    \label{fig:main_fig}
\end{figure}

With the onset of the digital age, millions of hours of video are being recorded and searching this data is becoming a monumental task. It is even more tedious when searching shifts from video-level labels, such as `dancing' or `skiing', to short action segments like `cracking eggs' or `tightening a screw'. In this paper, we focus on the latter and refer to them as fine-grained actions.
We thus explore the task of \textit{fine-grained action retrieval} where both queries and retrieved results can be either a video sequence, or a textual caption describing the fine-grained action. 
Such free-form action descriptions allow for a more subtle characterisation of actions but require going beyond training a classifier on a predefined set of action labels~\cite{mahdisoltani2018effectiveness, shao2018high}.

As is common in cross-modal search tasks~\cite{plummer2017enhancing,wang2016learning}, we learn a shared embedding space onto which we project both videos and captions.
By nature, fine-grained actions can be described by an actor, an act and the list of objects involved in the interaction.
We thus propose to learn a separate embedding for each part-of-speech (PoS), such as for instance verbs, nouns or adjectives.
This is illustrated in Fig.~\ref{fig:main_fig} for two PoS (verbs and nouns).
When embedding verbs solely, relevant entities are those that share the same verb/act regardless of the nouns/objects used. Conversely, for a PoS embedding focusing on nouns, different actions performed on the same object are considered relevant entities.
This enables a PoS-aware embedding, specialised for retrieving a variety of relevant entities, given that PoS.
The outputs from the multiple PoS embedding spaces are then combined within an encoding module that produces the final action embedding. We train our approach end-to-end, jointly optimising the multiple PoS embeddings and the final fine-grained action embedding.

This approach has a number of advantages over training a single embedding space as is standardly done~\cite{dong2018predicting, dong2018dual,hahn2018action,miech2018learning,mithun2018learning}.
Firstly, this process builds different embeddings that can be seen as different views of the data, which contribute to the final goal in a collaborative manner. 
Secondly, it allows to inject, in a principled way,
additional information but without requiring additional annotation, as parsing a caption for PoS is done automatically.
Finally, when considering a single PoS at a time, for instance verbs, the corresponding PoS-embedding learns to generalise across the variety of actions involving each verb (\eg the many ways `open' can be used). This generalisation is key to tackling more actions including new ones not seen during training.

We present the first retrieval results 
for the recent large-scale EPIC dataset~\cite{Damen2018EPICKITCHENS} (Sec~\ref{sec:experiments:epic}), utilising the released free-form narrations, previously unexplored for this dataset, as our supervision.
Additionally, we show that our second contribution, learning PoS-aware embeddings, is also valuable 
for general video retrieval by reporting results on the 
 MSR-VTT dataset~\cite{xu2016msr} (Sec.~\ref{sec:experiments:MSR}).

\section{Related Work} \label{sec:related_work}

Recently, neural networks trained with a 
 ranking loss  considering  image pairs~\cite{RadenovicECCV16CNN},  triplets~\cite{WangCVPR14Learning}, quadruplets~\cite{ChenCVPR17Beyond} or beyond~\cite{SohnNIPS16Improved}, 
have been considered for metric learning~\cite{HofferICLR15Deep,WangCVPR14Learning} and for a broad range of search tasks such as face/person identification~\cite{SchroffCVPR15FaceNet,ChenCVPR17Beyond,HermansX17Defense,almazan18reId} or instance retrieval~\cite{gordo2016deep,RadenovicECCV16CNN}.
These learning-to-rank approaches have been generalised to two or more modalities. Standard examples include building a joint embedding for images and 
text~\cite{gordo2017beyond,wang2016learning}, 
videos and audio~\cite{SurisECCVW18Cross-Modal} and, more related to our work, for videos and action labels~\cite{hahn2018action}, 
videos and text~\cite{dong2018dual,guadarrama2013youtube2text,xu2015jointly} or some of those combined~\cite{otani2016learning,mithun2018learning,miech2018learning}.

\myparagraph{Representing text}
Early works in image-to-text cross-modal retrieval~\cite{gong2014improving,gordo2017beyond,wang2016learning} used TF-IDF as a weighted bag-of-words model for text representations (either from a word embedding model or one-hot vectors) in order to aggregate variable length text captions into a single fixed sized representation. With the advent of neural networks, works shifted to use RNNs, Gated Recurrent Units (GRU) or Long Short-Term Memory (LSTM) units to extract textual features~\cite{dong2018dual} or to use these models within the embedding network~\cite{hahn2018action,kiros2014unifying,mithun2018learning,otani2016learning,torabi2016learning} for both modalities.

\myparagraph{Action embedding and retrieval}
Joint embedding spaces are a standard tool to perform action retrieval.
Zhang~\etal~\cite{zhang2015zero} use a Semantic Similarity Embedding (SSE) in order to perform action recognition.
Their method, inspired by sparse coding, splits train and test data into a mixture of proportions of already seen classes which then generalises to unseen classes at test time.
Mithun~\etal~\cite{mithun2018learning} create two embedding spaces: An activity space using flow and audio along with an object space from RGB.
Their method encodes the captions with GRUs and the output vectors from the activity and object spaces are concatenated to rank videos.
We instead create Part of Speech embedding spaces which we learn jointly, allowing our method to capture relationships between \eg verbs and nouns.

Hahn~\etal~\cite{hahn2018action} use two LSTMs to directly project videos into the Word2Vec embedding space. This method is evaluated on higher-level activities, showing that such a visual embedding aligns well with the learned space of Word2Vec to perform zero-shot recognition of these coarser-grained classes.
Miech~\etal~\cite{miech2017learnable} found that using NetVLAD~\cite{arandjelovic2016netvlad} results in an increase in accuracy over GRUs or LSTMs for aggregation of both visual and text features.
A follow up on this work~\cite{miech2018learning} learns a mixture of experts embedding from multiple modalities such as appearance, motion, audio or face features.
It learns a single output embedding which is the weighted similarity between the different implicit visual-text embeddings.
Recently, Miech~\etal~\cite{miech19howto100m} propose the HowTo100M dataset: A large dataset collected automatically using generated captions from youtube of `how to tasks'. They find that fine-tuning on these weakly-paired video clips allows for state-of-the-art performance on a number of different datasets.

\myparagraph{Fine-grained action recognition}
Recently, several large-scale datasets have been published for the task of fine-grained action recognition~\cite{Damen2018EPICKITCHENS,goyal2017_SomethingSomethingVideo,gu2018ava,sigurdsson2016hollywood,rohrbach2016recognizing}.
These generally focus on a closed vocabulary of class labels describing short and/or specific actions.

Rohrbach~\etal~\cite{rohrbach2016recognizing} investigate hand and pose estimation techniques for fine-grained activity recognition. By compositing separate actions, and treating them as attributes, they can predict unseen activities via novel combinations of seen actions.
Mahdisoltani~\etal~\cite{mahdisoltani2018effectiveness} train for four different tasks, including both coarse and fine grain action recognition. They conclude that training on fine-grain labels allows for better learning of features for coarse-grain tasks.

In our previous work~\cite{wray2019learning}, we explored action retrieval and recognition using multiple verb-only representations, collected via crowd-sourcing.
We found that a soft-assigned representation was beneficial for retrieval tasks over using the full verb-noun caption. 
While the approach enables scaling to a broader vocabulary of action labels, such multi-verb labels are expensive to collect for large datasets.

While focusing on fine-grained actions, we diverge from these works using open vocabulary captions for supervision. As recognition is not suitable, we instead formulate this as a retrieval problem. Up to our knowledge, no prior work attempted cross-modal retrieval on fine-grained actions. Our endeavour has been facilitated by the recent release of open vocabulary narrations on the EPIC dataset~\cite{Damen2018EPICKITCHENS} which we note is the only fine-grained dataset to do so. While our work is related to both \textit{fine-grained action recognition} and \textit{general action retrieval}, we emphasise that it is neither. We next describe our proposed model.

\section{Method} \label{sec:method}

\begin{figure*}
    \centering
    \includegraphics[width=\textwidth]{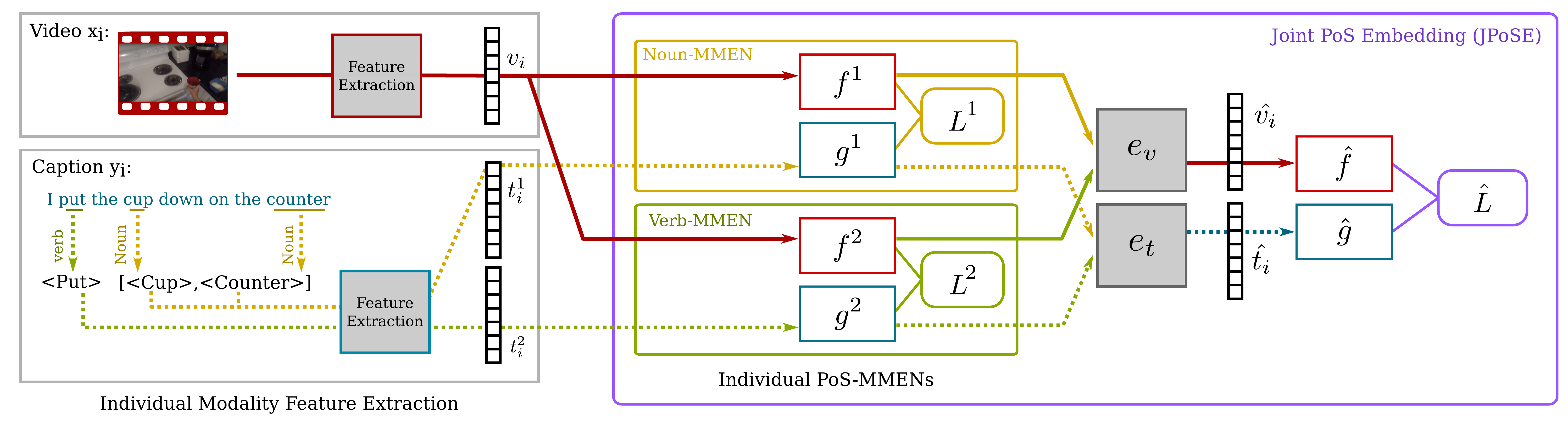}
    \caption{Overview of the \jp~model. We first disentangle a caption into its parts of speech (PoS) and learn a Multi-Modal Embedding Network (MMEN, Sec.~\ref{subsec:cross_modal}) for each PoS (Sec.~\ref{subsec:single_word_embedding}). The output of these PoS-MMENs are then encoded ($e_v$, $e_t$) to get new representations $\hat{v}_i$ and $\hat{t}_i$ on top of which the final embeddings $\hat{f}$ and $\hat{g}$ are learnt. \jp{} learns all of those jointly (Sec.~\ref{subsec:joint_meta_embedding}), using a combination of PoS-aware $L^1$, $L^2$, defined in Eq.~(\ref{eq:embed_loss}) and PoS-agnostic $\hat{L}$ losses, defined in Eq.~(\ref{eq:meta}). Non-trained modules are shown in grey.}
    \label{fig:method}
\end{figure*}

Our aim is to learn representations suitable for cross-modal search where the query modality is different from the target modality.
Specifically, we use video sequences with textual captions/descriptions and perform {\em video-to-text} (vt) or {\em text-to-video} (tv) retrieval tasks. 
Additionally, we would like to make sure that classical search (where the query and the retrieved results have the same modalities) could still be performed in that representation space. The latter are referred to as {\em video-to-video} (vv) and {\em text-to-text} (tt) search tasks.  
 As discussed in the previous section, several possibilities exist, the most common being embedding both modalities in a shared space such that, regardless of the modality, the representation of two relevant entities in that space are close to each other, while the representation of two non-relevant entities are far apart. 
  
 We first describe how to build such a joint embedding between two modalities, enforcing both cross-modal and within-modal constraints (Sec.~\ref{subsec:cross_modal}). Then, based on the knowledge that different parts of the caption encode different aspects of an action, we describe how to leverage this information and build several disentangled Part of Speech embeddings (Sec.~\ref{subsec:single_word_embedding}). Finally, we propose a unified representation well-suited for fine-grained action retrieval (Sec.~\ref{subsec:joint_meta_embedding}).

\subsection{Multi-Modal Embedding Network (\mm)} \label{subsec:cross_modal}

This section describes a Multi-Modal Embedding Network (MMEN) that encodes the video sequence and the 
text caption into a common descriptor space. 

Let $\{ (v_i,t_i) | v_i \in V, t_i \in T\}$ be a set of videos with $v_i$ being the visual representation of the $i^{th}$ video sequence and $t_i$ the corresponding textual caption.
Our aim is to learn two embedding functions $f: V  \rightarrow \Omega$ and  $g: T  \rightarrow \Omega$,  such that  $f(v_i)$ and $g(t_i)$ are close in the embedded space~$\Omega$. 
Note that $f$ and $g$ can be linear projection matrices or more complex functions \eg deep neural networks. We  denote  the parameters of the embedding functions $f$ and $g$  by $\theta_f$ and $\theta_g$ respectively,  
and we learn them jointly 
with a weighted combination of two cross-modal  ($L_{v,t},L_{t,v}$)
and two within-modal  ($L_{v,v},L_{t,t}$) triplet losses. Note that other point-wise, pairwise or
  list-wise losses can also be considered as alternatives to the triplet loss.

The \textbf{cross-modal losses} are crucial to the task and ensure that the representations of a query and a relevant item for that query from a different modality are closer than the representations of this query and a non-relevant item. We use cross-modal triplet losses \cite{wang2018learning, wang2016learning}: 
\begin{equation}
\begin{split}
    L_{v,t}(\theta)   =   \sum_{(i,j,k) \in \Tcal_{v,t}} \max \big(\gamma + d(f_{v_i}, g_{t_j}) - d(f_{v_i}, g_{t_k}), \, 0 \big) \\   
	\Tcal_{v,t} = \{(i,j,k) \, | \, v_i \in V,  t_j \in T_{i+}, t_k \in T_{i-}\} 
\end{split}
\label{eq:loss_vt}
\end{equation}
\begin{equation}
\begin{split}
    L_{t,v}(\theta) =  \sum_{(i,j,k) \in \Tcal_{t,v}} \max \big(\gamma + d( g_{t_i},f_{v_j}) - d(g_{t_i},f_{v_k}),\, 0 \big) \\
	\Tcal_{t,v} = \{(i,j,k) \, | \, t_i \in T,  v_j \in V_{i+}, v_k \in V_{i-}\} 
\end{split}
\label{eq:loss_tv}
\end{equation}
where $\gamma$ is a constant margin, $\theta = [\theta_f, \theta_g]$, and $d(.)$ is the distance function in the embedded space $\Omega$. $T_{i+},T_{i-}$ respectively define sets of relevant and non relevant captions and  $V_{i+},V_{i-}$ the sets of relevant and non relevant videos sequences for the multi-modal object $(v_i,t_i)$. To simplify the notation, $f_{v_i}$ denotes $f(v_i) \in \Omega$ and $g_{t_j}$ denotes $g(t_j)\in \Omega$. 

Additionally, \textbf{within-modal losses}, also called structure preserving losses~\cite{wang2018learning,wang2016learning}, ensure that the neighbourhood structure within each modality is preserved in the newly built joint embedding space. Formally,
\begin{equation}
\begin{split}
    L_{v,v}(\theta)   =   \sum_{(i,j,k) \in \Tcal_{v,v}} \max \big(\gamma + d(f_{v_i}, f_{v_j}) - d(f_{v_i}, f_{v_k}), \, 0\big) \\   
	\Tcal_{v,v} = \{(i,j,k) \, | \, v_i \in {V},  v_j \in V_{i+}, v_k \in V_{i-}\} 
\end{split}
\label{eq:loss_vv}
\end{equation}

\begin{equation}
\begin{split}
    L_{t,t}(\theta)   =   \sum_{(i,j,k) \in \Tcal_{t,t}} \max \big(\gamma + d(g_{t_i}, g_{t_j}) - d(g_{t_i}, g_{t_k}), \, 0 \big) \\   
	\Tcal_{t,t} = \{(i,j,k) \, | \, t_i \in T,  t_j \in T_{i+}, t_k \in T_{i-}\} 
\end{split}
\label{eq:loss_tt}
\end{equation}
using the same notation as before.
The final loss used for the \mm{} network is a weighted combination of these four losses, 
summed over all triplets in $\cal{T}$ defined as follows:
 \begin{equation}
     L (\theta)= \lambda_{v,v}L_{v,v} + \lambda_{v,t}L_{v,t} + \lambda_{t,v}L_{t,v} + \lambda_{t,t}L_{t,t}
    \label{eq:embed_loss}
\end{equation}
where $\lambda$ is a weighting for each loss term.

\subsection{Disentangled Part of Speech Embeddings} \label{subsec:single_word_embedding}

The previous section described the generic Multi-Modal Embedding Network (MMEN). 
In this section, we propose to disentangle different caption components so each component is encoded independently in its own embedding space. 
To do this, we first break down the text caption into different PoS tags. For example, the caption \textit{``I divided the onion into 
pieces using wooden spoon''} can be divided into verbs, $[divide,\,using]$, pronouns, $[I]$, nouns, $[onion, \, pieces, \, spoon]$ and adjectives, $[wooden]$. 
In our experiments, we focus on the most relevant ones for fine-grained action recognition: verbs and nouns, but we explore other types for general video retrieval.
We extract all words from a caption for a given PoS tag and train one \mm{} to only embed
these words and the video representation in the same space.
We refer to it as a PoS-MMEN.

To train a PoS-\mm, we propose to adapt the notion of relevance specifically to the PoS. This has a direct impact on the sets
$V_{i+},V_{i-},T_{i+},T_{i-}$ 
defined in Equations ({\ref{eq:loss_vt})-(\ref{eq:loss_tt}}).
For example, the caption \textit{`cut tomato'} is disentangled into the verb \textit{`cut'} and the noun \textit{`tomato'}.
Consider a PoS-\mm{} focusing on verb tags solely.
The caption \textit{`cut carrots'} is a relevant caption as the pair share the same verb \textit{`cut'}. In another PoS-\mm{} focusing on noun tags solely, the two remain irrelevant. As the relevant/irrelevant sets differ within each PoS-\mm{}, these embeddings specialise to that PoS.

It is important to note that, although the same visual features are used as input for all PoS-MMEN, the fact that we build one embedding space per PoS trains multiple visual embedding functions $f^k$ that can be seen as multiple views of the video sequence.


\subsection{PoS-Aware Unified Action Embedding} \label{subsec:joint_meta_embedding}

The previous section describes how to extract different PoS from captions and how to build PoS-specific MMENs. 
These PoS-MMENs can already be used alone for PoS-specific retrieval tasks, for instance a verb-retrieval task (\eg retrieve all videos where ``cut'' is relevant) or a noun-retrieval task.\footnote{Our experiments focus on action retrieval but we report on these other tasks in the supplementary material.} 
More importantly, the output of different PoS-MMENs can be combined to perform more complex tasks, including the one we are interested in, namely fine-grained action retrieval.

Let us denote the $k^{\text{th}}$ PoS-MMEN visual and textual embedding functions by $f^k: V  \rightarrow \Omega^k$ and $g^k: T  \rightarrow \Omega^k$. We define:
\begin{equation}
\begin{split}
\hat{v}_i=e_v(f^1_{v_i},f^2_{v_i},\ldots, f^K_{v_i}) \\ \hat{t}_i=e_t(g^1_{t^1_i},g^2_{t^2_i},\ldots,g^K_{t^K_i})
\end{split}
\label{eq:meta}
\end{equation}
where  $e_v$ and $e_t$ are encoding functions that combine the outputs of the PoS-MMENs.
We explore multiple pooling functions for $e_v$ and $e_t$: \textit{concatenation}, \textit{max}, \textit{average} - the latter two assume all $\Omega_k$ share the same dimensionality.

When $\hat{v}_i$, $\hat{t}_i$ have the same dimension, we can perform action retrieval by directly computing the distance between these representations. We instead propose to train a final PoS-agnostic MMEN that unifies the representation, leading to our final \jp~model.

\myparagraph{Joint Part of Speech Embedding (\jp)}
Considering the PoS-aware representations  $\hat{v}_i$ and $\hat{t}_i$ as input and, still following our learning to rank approach, we learn the parameters ${\hat{\theta}_{\hat{f}}}$ and $\hat{\theta}_{\hat{g}}$ of the two embedding functions ${\hat{f}: \hat{V} \rightarrow  \Gamma}$ and 
$\hat{g}: \hat{T} \rightarrow  \Gamma$ which project in our final embedding space $\Gamma$.
We again consider this as the task of building a single MMEN with the inputs $\hat{v}_i$ and $\hat{t}_i$, and follow the process described in Sec. \ref{subsec:cross_modal}. In other words, we train using the loss defined in Equation~(\ref{eq:embed_loss}), which we denote  $\hat{L}$ here, which combines two cross-modal and two within-modal losses using the triplets $\Tcal_{v,t},\Tcal_{t,v},\Tcal_{v,v},\Tcal_{t,t}$ formed using relevance between videos and captions 
based on the action retrieval task. As relevance here is not PoS-aware, we refer to this loss as PoS-agnostic.
This is illustrated in Fig.~\ref{fig:method}.

We
learn the multiple PoS-\mm{}s and the final \mm{} jointly with the following combined loss:
\begin{equation}
    L(\hat{\theta},\theta^1,\ldots \theta^K) =  \hat{L}(\hat{\theta}) +  \sum^K_{k=1} \alpha^k  L^k(\theta^k)
\label{eq:JPME}
 \end{equation}
 where $\alpha^k$ are weighting factors, $\hat{L}$ is the PoS-agnostic loss described above and  $L^k$ are the PoS-aware losses corresponding to the $K$ PoS-MMENs.

\section{Experiments}
\label{sec:experiments}

We first tackle fine-grained action retrieval on the EPIC dataset~\cite{Damen2018EPICKITCHENS} (Sec.~\ref{sec:experiments:epic}) and then the general video retrieval task on the MSR-VTT dataset~\cite{xu2016msr} (Sec.~\ref{sec:experiments:MSR}).
This allows us to explore two different tasks using the proposed multi-modal embeddings. 

The large English spaCy parser~\cite{spacy} was used to find 
the Part Of Speech (PoS) tags and disentangle them in the captions of both datasets. Statistics on the most frequent PoS tags are shown in Table~\ref{tab:pos}.
As these statistics show, EPIC contains mainly nouns and verbs, while  MSR-VTT has longer captions and more nouns. This will have an impact of the PoS chosen for each dataset when building the \jp{} model.

\subsection{Fine-Grained Action Retrieval on EPIC}
\label{sec:experiments:epic}

\myparagraph{Dataset}
The EPIC dataset~\cite{Damen2018EPICKITCHENS} is an egocentric dataset with 32 participants cooking in their own kitchens who then narrated the actions in their native language.
The narrations were translated to English but maintain the open vocabulary selected by the participants.
We employ the released free-form narrations to use this dataset for fine-grained action retrieval.
We follow the provided train/test splits. Note that by construction there are two test sets: \textit{Seen} and \textit{Unseen}, referring to whether the kitchen has been seen in the training set. We follow the terminology from~\cite{Damen2018EPICKITCHENS}, and note that this terminology should not be confused with the zero-shot literature which distinguishes seen/unseen classes. 
The actual sequences are strictly disjoint between all sets.

Additionally, we train only on the many-shot examples from EPIC excluding all examples of the few shot classes from the training set. This ensures each action has more than 100 relevant videos during training and increases the number of zero-shot examples in both test sets.

\myparagraph{Building relevance sets for retrieval}
The EPIC dataset offers an opportunity for fine-grained action retrieval, as the open vocabulary has been grouped into semantically relevant verb and noun classes for the action recognition challenge.
For example, {\em `put'}, {\em `place'} and {\em `put-down'} are grouped into one class.
As far as we are aware, this paper presents the first attempt to use the open vocabulary narrations 
released to the community.
We determine retrieval relevance scores from these semantically grouped verb and noun classes\footnote{We use the verb and noun classes purely to establish relevance scores, the training is done with the original open vocabulary captions.}, defined in~\cite{Damen2018EPICKITCHENS}.
These indicate which videos and captions should be considered related to each other.
Following these semantic groups, a query {\em `put mug'} and a  video with {\em `place cup'} in its caption are considered relevant as {\em `place'} and {\em `put'} share the same verb class and 
{\em `mug'} and {\em `cup'} share the same noun class.
Subsequently, we define the triplets $\Tcal_{v,t},\Tcal_{t,v},\Tcal_{v,v},\Tcal_{t,t}$  used to train the \mm{} models and to compute the loss $\hat{L}$  in \jp.

When training a PoS-MMEN, two videos are considered relevant only within that PoS. Accordingly, {\em`put onion'} and {\em`put mug'} are relevant for verb retrieval, whereas, {\em`put cup'} and {\em`take mug'} are for noun retrieval. The corresponding PoS-based relevances define the triplets  
$\Tcal^k$ for $L^k$.

\begin{table}[t]
\footnotesize
    \centering
    \begin{tabularx}{\linewidth}{lcrYcYY}
    \toprule
                    & & \multicolumn{2}{c}{EPIC} & & \multicolumn{2}{c}{MSR-VTT}\\
    Parts of Speech & & count & avg/caption & & count & avg/caption \\ \midrule
    Noun           & & 34,546 & 1.21 & & 418,557 &  3.33 \\
    Verb           & & 30,279 & 1.06 & & 245,177 &  1.95 \\
    Determiner     & &  6,149 & 0.22 & & 213,065 &  1.69 \\
    Adposition     & &  5,048 & 0.18 & & 151,310 &  1.20 \\
    Adjective      & &  2,271 & 0.08 & &  79,417&  0.63 \\
    \bottomrule \\
    \end{tabularx}
    \caption{Statistics of the 5 most common PoS tags in the training sets of both datasets: total counts and average counts per caption. }
    \label{tab:pos}
\end{table}

\subsubsection{Experimental Details}

\myparagraph{Video features} 
We extract flow and appearance features using the TSN BNInception model \cite{TSN2016ECCV}  pre-trained on Kinetics and  fine-tuned on our training set.  TSN averages the features from 25 uniformly sampled snippets within the video. We then concatenate appearance and flow features to create a 2048 dimensional vector ($v_i$) per action segment.

\myparagraph{Text features} 
We map each lemmatised word to its feature vector using a 100-dimension Word2Vec model, trained on the Wikipedia corpus. Multiple word vectors with the same part of speech were aggregated by averaging. 
We also experimented with the pre-trained 300-dimension Glove model, and found the results to be similar.

\myparagraph{Architecture details}
 We implement $f^k$ and $g^k$ in each \mm{}
as a 2 layer perceptron  (fully connected layers) with ReLU.  
Additionally, the input vectors and output vectors are L2 normalised. In all cases, we set the dimension of the embedding space to 256, a dimension we found to be suitable across all  settings.
We use a single layer perceptron with shared weights for $\hat{f}$ and $\hat{g}$ that we initialise with PCA.

\myparagraph{Training details}
The triplet weighting parameters 
are set to $\lambda_{v,v}=\lambda_{t,t}=0.1$ and $\lambda_{v,t}=\lambda_{t,v}=1.0$ and the loss weightings $\alpha^k$ are set to $1$.
The embedding models were implemented in Python using the Tensorflow library. We trained the models with an Adam solver and a learning rate of $1e^{-5}$, considering  batch sizes of 256, where for each query we sample 100 random triplets from the corresponding $\Tcal_{v,t},\Tcal_{t,v},\Tcal_{v,v},\Tcal_{t,t}$ sets.
The training in general converges after a few thousand iterations, we report all results after 4000 iterations. 

\myparagraph{Evaluation metrics}
We report mean average precision~(mAP), \ie for each query we consider the average precision over all relevant elements and take the mean over all queries.
We consider each element in the test set as a query in turns.
When reporting within-modal retrieval mAP, the corresponding item (video or caption) is removed from the test set for that query.

\subsubsection{Results}

First, we consider  cross-modal  and within-modal fine-grained action retrieval. Then, we present an ablation study as well as qualitative results to get more insights. Finally we show that our approach is well-suited for zero-shot settings.

\myparagraph{Compared approaches} Across a wide of range of experiments, we compare the proposed \textbf{\jp} (Sec.~\ref{subsec:joint_meta_embedding}) with some simpler variants based on \textbf{MMEN} (Sec. ~\ref{subsec:cross_modal}).

For the captions, we use 1) all the words together without distinction, denoted as `Caption',  2) only one PoS such as `Verb' or `Noun', 3) the concatenation of their respective representations, denoted as `[Verb, Noun]'.

These models are also compared to standard baselines. The \textbf{Random Baseline} randomly ranks all the database items, providing a lower bound on the mAP scores. The {\textbf{CCA-baseline}} applies Canonical Correlation Analysis to both modalities $v_i$ and $t_i$ to find a joint embedding space for cross-modal retrieval~\cite{gong2014improving}.
Finally, \textbf{Features~(Word2Vec)} and \textbf{Features~(Video)}, which are only defined for within-modal retrieval (\ie vv and tt), show the performance when we directly use the video representation $v_i$ or the averaged  Word2Vec caption representation $t_i$.

\begin{table}[]
\centering
\footnotesize
    \begin{tabular}{lcrrcrr}
    \toprule
                      EPIC  & & \multicolumn{2}{c}{SEEN} & & \multicolumn{2}{c}{UNSEEN}\\
                            & & vt & tv & & vt & tv  \\ \midrule
    Random Baseline         & &0.6 &0.6 & &0.9 &0.9  \\
    CCA Baseline            & &20.6&7.3 & &14.3&3.7  \\ \midrule
    \mm~(Verb)              & &3.6 &4.0 & &3.9 &4.2  \\
    \mm~(Noun)              & &9.9 &9.2 & &7.9 &6.1  \\
    \mm~(Caption)           & &14.0&11.2& &10.1&7.7  \\
    \mm~([Verb, Noun])      & &18.7&13.6& &13.3&9.5  \\ \midrule
    \jp~(Verb, Noun)         & &\textbf{23.2}& \textbf{15.8}& & \textbf{14.6}& \textbf{10.2} \\
    \bottomrule \\
    \end{tabular}
    \caption{Cross-modal action retrieval on EPIC.}
    \label{tab:cross_modal_results}
\end{table}

\begin{table}[]
\centering
\footnotesize
    \begin{tabular}{lcrrcrr}
    \toprule
    EPIC                    & & \multicolumn{2}{c}{SEEN} & & \multicolumn{2}{c}{UNSEEN}\\
                            & & vv & tt & & vv & tt  \\ \midrule
    Random Baseline         & & 0.6& 0.6& & 0.9&0.9  \\
    CCA Baseline            & &13.8&62.2& &18.9&68.5 \\
    Features~(Word2Vec)      & & -- &62.5& & -- &71.3 \\
    Features~(Video)         & &13.6& -- & &21.0& --  \\ \midrule
    \mm~(Verb)              & &15.2&11.7& &20.1&15.8 \\
    \mm~(Noun)              & &16.8&30.1& &21.2&34.1 \\
    \mm~(Caption)           & &17.2&63.8& &20.7&69.6 \\
    \mm~([Verb, Noun])      & &17.6&83.5& &22.5&84.7 \\ \midrule
    \jp~(Verb, Noun)         & &\textbf{18.8}&\textbf{87.7}& &\textbf{23.2}&\textbf{87.7} \\ 
    \bottomrule \\
    \end{tabular}
    \caption{Within-modal action retrieval on EPIC.}
    \label{tab:within_modal_results}
\end{table}

\myparagraph{Cross-modal retrieval} Table~\ref{tab:cross_modal_results}  presents cross-modal results for fine-grained action retrieval.
The main observation is that the proposed \jp{} outperforms all the \mm{} variants and the baselines for both video-to-text  (vt) and text-to-video retrieval (tv), on both test sets.
We also note that \mm~([Verb, Noun]) outperforms other \mm{} variants, showing the benefit of learning specialised embeddings. Yet the full \jp{} is crucial to get the best results.

\myparagraph{Within-modal retrieval}
Table~\ref{tab:within_modal_results} shows the within-modal retrieval results for 
both text-to-text~(tt) and video-to-video~(vv) retrieval. Again, \jp{} outperforms all the flavours of \mm{} on both test sets.
This shows that by learning a cross-modal embedding we inject information from the other modality that helps to better disambiguate and hence to improve the search.

\begin{figure*} 
    \centering
    \includegraphics[width=\linewidth]{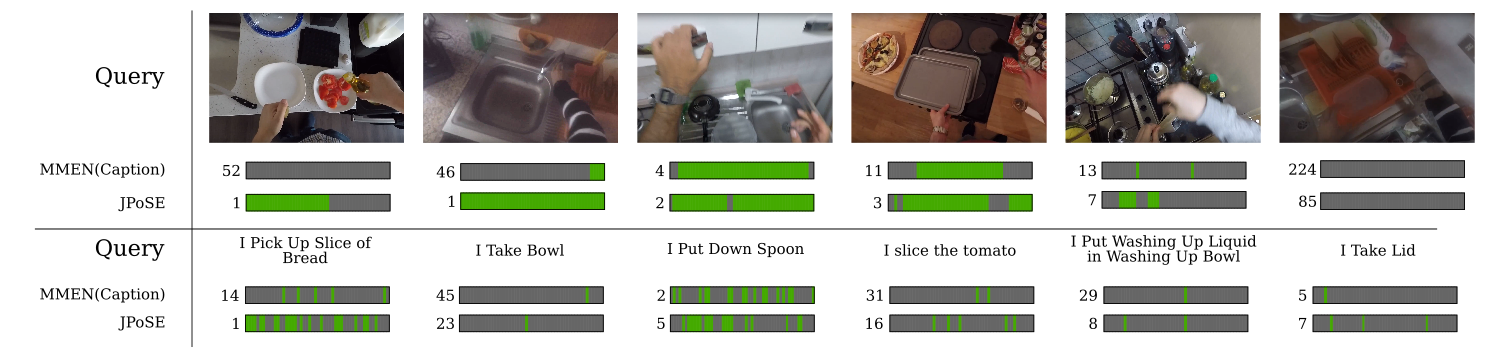}
    \caption{Qualitative results for video-to-text (top) and text-to-video (bottom) action retrieval on EPIC. For several query videos (top) or query captions (bottom), we show the quality of the top 50 captions (resp. videos) retrieved with green/grey representing relevant/irrelevant retrievals. The number in front of the colour-coded bar shows the rank of the first relevant retrieval (lower rank is better).}
    \label{fig:qualitative_tv_vt}
\end{figure*}

\begin{figure*}
    \centering
    \includegraphics[width=\linewidth]{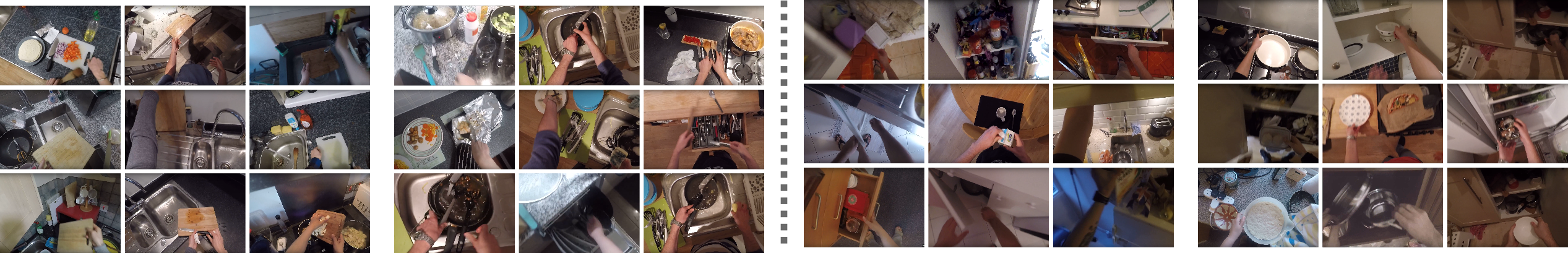}
    \caption{Maximum activation examples for visual embedding in the noun (left) and the verb (right) \ip. Examples of similar objects over different actions are shown in the noun embedding (left) [chopping board vs cutlery] while the same action is shown over different objects in the verb embedding (right) [open/close vs put/take].}
    \label{fig:spaces_qualitative}
\end{figure*}

\myparagraph{Ablation study}
We evaluate the role of the components of the proposed \jp{} model, for both cross-modal and within-modal retrieval. Table~\ref{tab:epic_ablation} reports results comparing different options for the encoding functions $e_{v}$ and $e_{t}$ in addition to learning the model jointly both with and without learned functions $\hat{f}$ and $\hat{g}$.  This confirms that the proposed approach 
is the best option.
In the supplementary material, we also compare the performance when using the closed vocabulary classes from EPIC to learn the embedding. Results demonstrate the benefit of utilising the open vocabulary at training time.

\begin{table}[]
\footnotesize
    \centering
    \begin{tabular}{ccccrrrrr}
    \toprule
         EPIC     & & & & & \multicolumn{4}{c}{SEEN}\\
  Learn & $\hat{L}$    &($e_v$,$e_t$)&($\hat{f}, \hat{g}$)& & vv & vt & tv & tt \\ \midrule
  indep & $\times$     &   Sum       & $\times$           & &17.4&20.7&13.3&86.5\\
  indep & $\times$     &   Max       & $\times$           & &17.5&21.2&13.3&86.5\\
  indep & $\times$     &   Conc.     & $\times$           & &18.3&21.5&14.6&87.1\\ \midrule
  joint & $\checkmark$ &   Sum       & $(Id,Id)$          & &18.1&21.0&14.3&87.3\\
  joint & $\checkmark$ &   Max       & $(Id,Id)$          & &18.1&22.4&14.8&87.5\\
  joint & $\checkmark$ &   Conc.     & $(Id,Id)$          & &18.3&22.7&15.4&87.6\\ \midrule
  joint & $\checkmark$ &   Conc.     & $(\hat{\theta}_{\hat{f}} ,\hat{\theta}_{\hat{g}})$ & &\textbf{18.8}&\textbf{23.2}&\textbf{15.8}&\textbf{87.7}\\
    \bottomrule \\    
    \end{tabular}    
    \caption{Ablation study for \jp{} showing the effects of different encodings,  training PoS-\mm{}s, independently or jointly with $\hat{f}$ and $\hat{g}$ being the identity function $Id$ or being learnt.}
    \label{tab:epic_ablation}
\end{table}

\myparagraph{Zero-shot experiments}
The use of the open vocabulary in EPIC lends itself well to zero-shot settings. These are the cases for which the open vocabulary verb or noun in the test set is not present in the training set.  Accordingly, all previous results can be seen as a Generalised Zero-Shot Learning (GZSL)~\cite{chao2016empirical} set-up: there exists both actions in the test sets that have been seen in training and actions that have not. Table~\ref{tab:epic_vn_counts} shows the zero-shot (ZS) counts in both test sets. In total 12\% of the videos in both test sets are zero-shot instances.
We separate cases where the noun is present in the training set but the verb is not, denoted by ZSV (zero-shot verb), from ZSN (zero-shot noun) where the verb is present but not the noun.  Cross-modal ZS retrieval results for this interesting setting are shown in Table~\ref{tab:zs_results}.
We compare \jp{} to \mm~(Caption) and baselines.  

Results show that the proposed \jp{} model clearly improves over these zero-shot settings, thanks to the different views captured by the multiple PoS embeddings, specialised to acts and objects.

\begin{table}[]
\footnotesize
    \centering
    \begin{adjustbox}{max width=\linewidth}
    \begin{tabular}{lrrrrrrr}
    \toprule
    EPIC              & \multicolumn{3}{c}{All}     & \multicolumn{2}{c}{ZSV}       & \multicolumn{2}{c}{ZSN} \\
                      &  Videos  &  Verbs &  Nouns  & Videos  &  Verbs    & Videos  & Nouns \\\midrule
    Train             &  26,710  & 192    & 1005    & --      &  --       &  --     &  --   \\
    Seen Test              &   8,047  & 232    &  530    & 452      & 119       & 367     &  80   \\
    Unseen Test           &   2,929  & 136    &  243    & 257      &  63       & 275     & 127   \\
    \bottomrule \\
    \end{tabular}
    \end{adjustbox}
    \caption{Number of videos, and number of open-vocabulary verbs and nouns in the captions, for the three splits of EPIC. For both test sets we also report zero-shot (ZS) instances, showing the number of verbs and nouns that were not seen in the training set, as well as the corresponding number of videos.}
    \label{tab:epic_vn_counts}
\end{table}

\begin{table}[]
\footnotesize
    \centering
    \begin{tabularx}{.85\linewidth}{lYYYY}
    \toprule
                         EPIC &\multicolumn{2}{c}{ZSV} & \multicolumn{2}{c}{ZSN} \\
                         & vt & tv & vt & tv  \\ \midrule
    Random Baseline      &1.57&1.57&1.64&1.64 \\
    CCA Baseline         &2.92&2.96&4.36&3.25 \\ \midrule
    \mm~(Caption)    &5.77&5.51&4.17&3.32 \\
    \jp                  &{\bf7.50}&{\bf6.47}&{\bf7.68}&{\bf6.17} \\
    \bottomrule \\
    \end{tabularx}
    \vspace*{-6pt}
    \caption{Zero shot experiments on EPIC.}
    \label{tab:zs_results}
\end{table}

\myparagraph{Qualitative results}
Fig.~\ref{fig:qualitative_tv_vt} illustrates both video-to-text and text-to-video retrieval.
For several queries, it shows the relevance of the top-50 retrieved items (relevant in green, non-relevant in grey).

Fig.~\ref{fig:spaces_qualitative} illustrates our motivation that disentangling PoS embeddings would learn different visual functions. It presents maximum activation examples on chosen neurons within $f_i$ for both verb and noun embeddings. Each cluster represents the 9 videos that respond maximally to one of these neurons\footnote{Video can be accessed at \url{https://www.youtube.com/watch?v=FLSlRQBFow0}.}.
We can remark that noun activations indeed correspond to objects of shared appearance occurring in different actions (in the figure, chopping boards in one and cutlery in the second), while verb embedding neuron activations reflect the same action applied to different objects (open/close \vs put/take).

\subsection{General Video Retrieval on MSR-VTT}
\label{sec:experiments:MSR}

\begin{table*}[t]
\footnotesize
    \centering
    \begin{tabularx}{.9\linewidth}{lcYYYYcYYYY}
    \toprule
                           & & \multicolumn{4}{c}{Video-to-text} & & \multicolumn{4}{c}{Text-to-Video} \\
    MSR-VTT Retrieval      & &  R@1 &  R@5 & R@10 &  MR    & &  R@1 &  R@5 & R@10 &  MR  \\ \midrule
    Mixture of Experts~\cite{miech2018learning}*& &  --    &  --  &  --  &  --  & & 12.9 & 36.4 & 51.8 & 10.0   \\ \midrule
    Random Baseline        & & 0.3 & 0.7 & 1.1 & 502.0    & & 0.3 & 0.7 & 1.1 & 502.0    \\
    CCA Baseline           & & 2.8 & 5.6 & 8.2 & 283.0    & &  7.0 & 14.4 & 18.7 & 100.0    \\ \midrule
    \mm~(Verb)      & & 0.7 & 4.0 & 8.3 &  70.0    & &  2.9 &  7.9 & 13.9 &  63.0    \\
    \mm~(Caption$\setminus$Noun) & & 5.7  & 18.7 & 28.2 & 31.1  & & 5.3  & 17.0 & 26.1 & 33.3   \\
    \mm~(Noun)      & & 10.8 & 31.3 & 42.7 &  14.0    & & 10.8 & 30.7 & 44.5 &  13.0    \\
    \mm~([Verb, Noun]) & & 15.6 & 39.4 & {\bf 55.1} &   9.0    & & 13.6 & 36.8 & 51.7 &  10.0    \\ 
    \mm~(Caption)   & & 15.8 & 40.2 & 53.6 &   9.0   & & 13.8 & 36.7 & 50.7 &  10.3  \\  \midrule 
    \jp~(Caption$\setminus$Noun, Noun)                   & & {\bf16.4} & {\bf41.3} & 54.4 &   {\bf8.7}  & & {\bf14.3} & {\bf38.1} & {\bf53.0} &   {\bf9.0}    \\
    \bottomrule \\
    \end{tabularx}
    \caption{MSR-VTT Video-Caption Retrieval results. *We include results from ~\cite{miech2018learning}, only available for Text-to-Video retrieval.} 
    \label{tab:msr_vtt}
\end{table*}
\begin{figure*}
    \centering
    \includegraphics[width=\linewidth]{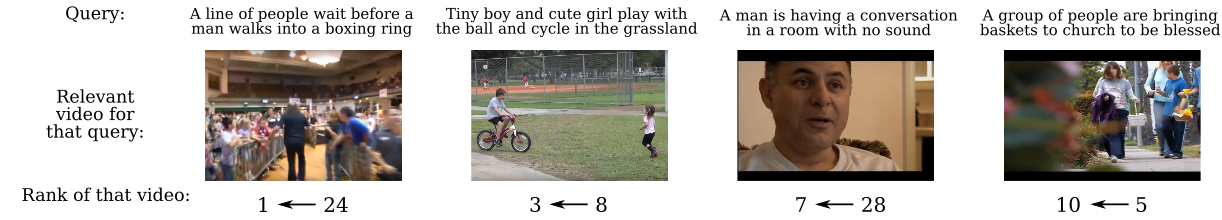}
    \caption{Qualitative results of text-to-video action retrieval on MSR-VTT. $A \leftarrow B$ shows the rank $B$ of the retrieved video from using the full caption \mm~(caption) and the rank $A$ when disentangling the caption \jp~(Caption$\setminus$Noun, Noun).}
    \label{fig:qual_msr}
\end{figure*}


\myparagraph{Dataset} We select {\bf MSR-VTT~\cite{xu2016msr}} as a public dataset for general video retrieval. Originally used for video captioning, this large-scale video understanding dataset is increasingly evaluated for video-to-text and text-to-video retrieval~\cite{dong2018dual,miech2018learning,mithun2018learning,yu2018joint,miech19howto100m}.
We follow the code and setup of~\cite{miech2018learning} using the same train/test split that includes 
7,656 training videos each with 20 different captions describing the scene and 1000 test videos 
with one caption per video. 
We also follow the evaluation protocol in~\cite{miech2018learning} and compute recall@k (R@K) and median rank (MR).

In contrast to the EPIC dataset, there is no semantic groupings of the captions in MSR-VTT. Each caption is considered relevant only for a single video, and two captions 
describing different videos are considered irrelevant even if they share semantic similarities. Furthermore, disentangling captions yields further semantic similarities. For example, ``\textit{A cooking tutorial}''  and ``\textit{A person is cooking}'', for a verb-\mm, will be considered irrelevant as they belong to different videos even though they share the same single verb `cook'.

Consequently, we can not directly apply \jp{} as proposed in Sec.~\ref{subsec:joint_meta_embedding}. Instead, we adapt  
 \jp{} to this problem as follows.
 We use the Mixture-of-Expert Embeddings (MEE) model from~\cite{miech2018learning}, as our core MMEN network. In fact, MEE is a form of multi-modal embedding network in that it embeds videos and captions into the same space. We instead focus on assessing whether disentangling PoS and learning multiple PoS-aware embeddings produce better results.
 In this adapted \jp{} we encode the output of the  disentangled PoS-\mm{}s with $e_v$ and $e_t$ (\ie concatenated) and use NetVLAD~\cite{arandjelovic2016netvlad} to aggregate Word2Vec representations.  
 Instead of the combined loss in Equation (\ref{eq:JPME}), 
 we  use  the pair loss, used also in \cite{miech2018learning}:
 \begin{equation}
\begin{split}
    L(\theta)   =   \frac{1}{B} \sum_i^B \sum_{j \neq i} \max \big(\gamma + d(f_{v_i}, g_{t_i}) - d(f_{v_i}, g_{t_j}), \, 0 \big) \\ + \max \big(\gamma + d(f_{v_i}, g_{t_i}) - d(f_{v_j}, g_{t_i}) , \, 0 \big)   
\end{split}
\label{eq:loss_meee}
 \end{equation}
 This same loss is used when we train different \mm{}s. 

\myparagraph{Visual and text features}
We use appearance, flow, audio and facial pre-extracted visual features provided from~\cite{miech2018learning}. For the captions, we extract the encodings ourselves\footnote{Note that this explains the difference between the results reported in~\cite{miech2018learning} (shown in the first row of the Table \ref{tab:msr_vtt}) and MMEN (Caption).}
using the same Word2Vec model as for EPIC. 

\myparagraph{Results}
We report on video-to-text and text-to-video retrieval on MSR-VTT in Table~\ref{tab:msr_vtt} for the standard baselines and several \mm{} variants. 
Comparing \mm{}s, we note that nouns are much more informative than verbs for this retrieval task.
MMEN results with other PoS tags (shown in the supplementary) are even lower, indicating that they are not informative alone.
Building on these findings, we report results of a JPoSE combining two \mm{}s, one for nouns, and one for the remainder of the caption (Caption$\setminus$Noun). 
 Our adapted \jp{} model consistently outperforms full-caption single embedding for both video-to-text and text-to-video retrieval.
 We report other PoS disentanglement results in supplementary material.

\myparagraph{Qualitative results} Figure~\ref{fig:qual_msr} shows qualitative results comparing using the full caption and \jp{} noting the disentangled model's ability to commonly rank videos closer to their corresponding captions.



\section{Conclusion} \label{sec:conclusion}

We have proposed a method for fine-grained action retrieval. By learning distinct embeddings for each PoS, our model is able to combine these in a principal manner and to create a space suitable for action retrieval, outperforming approaches which learn such a space through captions alone.
We tested our method on a fine-grained action retrieval dataset, EPIC, using the open vocabulary labels. Our results demonstrate the ability for the method to generalise to zero-shot cases.
Additionally, we show the applicability of the notion of disentangling the caption for the general video-retrieval task on MSR-VTT.

\noindent \textbf{Acknowledgement} \hspace{6pt} Research supported by EPSRC LOCATE (EP/N033779/1) and {EPSRC} Doctoral Training Partnerships (DTP). We use publicly available datasets. Part of this work was carried out during Michael's internship at Naver Labs Europe.

\noindent\textbf{Project Page} \hspace{6pt} The project page for this work can be found at \url{https://mwray.github.io/FGAR}.

{\small
\bibliographystyle{ieee_fullname}
\bibliography{egbib}
}

\newpage

{
    \huge
\section*{Supplementary Material}
}

\appendix
\section{Individual Part-of-Speech Retrieval \newline (Sec.~3.3)}

In the main manuscript, we report results on the task of \textit{fine-grained action retrieval}.
For completion, we here present results on individual Part-of-Speech~(PoS) retrieval tasks.

In Table~\ref{tab:unseen_verb}, we report results for  \textit{fine-grained verb retrieval} (\ie only retrieve the relevant verb/action in the video). 
We include the standard baselines and we additionally report the results obtained by a PoS-MMEN, that is a single embedding for verbs solely. We compare this to our proposed multi-embedding \jp{}.
Using \jp{} produces better (or the same) results for both cross-modal and within-modal searches.

Similarly, in Table~\ref{tab:unseen_noun}, we compare results for \textit{fine-grained noun retrieval} (\ie only retrieve the relevant noun/object in the video). We show similar increases in mAP over cross-modal and within-modal searches. This indicates the complementary PoS information, from the other PoS embedding as well as the PoS-aware action embedding, helps to better define the individual embedding space.

\begin{table}[t]
\footnotesize
    \centering
    \begin{tabularx}{\linewidth}{lYYYY}
    \toprule
                        EPIC& \multicolumn{4}{c}{SEEN} \\
                        & vv & vt & tv & tt \\ \midrule
    Random Baseline     &12.6&12.6&12.6&12.6\\ 
    Features(Word2Vec)  & -- & -- & -- &50.0\\ 
    Features(Video)     &21.0& -- & -- & -- \\ 
    CCA Baseline        &21.3&23.3&25.7&37.7\\ \midrule
    \mm(Caption)        &32.0&53.1&47.2&90.0\\ 
    \mm([Verb,Noun])    &33.2&55.7&48.9&96.1\\ 
    \mm(Caption RNN)    &31.2&33.7&49.2&92.6\\ \midrule
    PoS-\mm(Verb)       &31.1&56.2&48.5&{\bf97.1}\\ \midrule
    \jp                 &{\bf33.7}&{\bf57.1}&{\bf49.9}&{\bf97.1}\\ 
    \bottomrule \\
    \end{tabularx}
    \caption{Verb retrieval task results on the seen test set of EPIC-Kitchens.}
    \label{tab:unseen_verb}
\end{table}

\begin{table}[]
\footnotesize
    \centering
    \begin{tabularx}{\linewidth}{lYYYY}
    \toprule
                        EPIC& \multicolumn{4}{c}{SEEN} \\
                        & vv & vt & tv & tt  \\ \midrule
    Random Baseline     &2.17&2.17&2.17&2.17\\ 
    Features(Word2Vec)  & -- & -- & -- &30.9\\ 
    Features(Video)     &10.6& -- & -- & -- \\ 
    CCA Baseline        &11.9&16.9&19.2&52.2\\ \midrule
    \mm(Caption)        &{\bf18.7}&26.2&20.7&70.9\\ 
    \mm([verb,Noun])    &18.3&29.8&23.8&90.1\\ 
    \mm(Caption RNN)    &17.9&20.3&22.0&74.0\\ \midrule
    PoS-\mm(Noun)       &17.8&31.5&23.6&{\bf92.6}\\ \midrule
    \jp                 &18.6&{\bf32.2}&{\bf25.5}&{\bf92.6}\\ 
    \bottomrule
    \end{tabularx}
    \caption{Noun retrieval task results on the seen test set of EPIC-Kitchens.}
    \label{tab:unseen_noun}
\end{table}

\section{Closed vs Open Vocabulary Embedding}
Table~\ref{tab:cross_modal_results_closed} 
compares to \jp{}* trained using only the closed vocabulary of EPIC. 
In this setup, closed vocabulary was used for building the embedding, but open vocabulary used for testing.
Results show that using the full open vocabulary in training yields a sizeable benefit.

\begin{table}[t]
\footnotesize
 \centering
    \begin{tabular}{lcrrcrr}
    \toprule
                           EPIC  & & \multicolumn{2}{c}{SEEN} & & \multicolumn{2}{c}{UNSEEN}\\
                               & & vt & tv & & vt & tv  \\ \midrule
    \jp (Verb,Noun)*             & &18.0&13.4& &11.5&8.8  \\ 
    \jp (Verb,Noun)             & &\textbf{23.2}& \textbf{15.8}& & \textbf{14.6}& \textbf{10.2} \\
    \bottomrule \\
    \end{tabular}
    \caption{Cross-modal retrieval results - compared closed~(*) to open vocabulary embedding.}
    \label{tab:cross_modal_results_closed}
\end{table}
\vspace*{-5pt}

\section{Text embedding Using RNN}
We provide here the results of replacing the text embedding function, $g$, with an RNN instead of the two layer perceptron for the \mm{} method.
The RNN was modelled as a Gated Recurrent Unit (GRU). Captions were capped and zero-padded to a maximum length  of $15$ words. Adding a layer on top of the GRU proved not to be useful. Results of the RNN in the experiments are given under the name MMEN~(Caption RNN). Given the singular verb and low noun count RNNs were not tested for the individual \ip s.

\begin{table}[]
\centering
\footnotesize
    \begin{tabular}{lcrrcrr}
    \toprule
                      EPIC  & & \multicolumn{2}{c}{SEEN} & & \multicolumn{2}{c}{UNSEEN}\\
                            & & vt & tv & & vt & tv  \\ \midrule
    Random Baseline         & &0.6 &0.6 & &0.9 &0.9  \\
    CCA Baseline            & &20.6&7.3 & &14.3&3.7  \\ \midrule
    \mm~(Caption)           & &14.0&11.2& &10.1&7.7  \\
    \mm~(Caption RNN)       & &10.3&13.8& &6.3 &9.0  \\
    \mm~([Verb, Noun])      & &18.7&13.6& &13.3&9.5  \\ \midrule
    \jp (Verb,Noun)         & &\textbf{23.2}& \textbf{15.8}& & \textbf{14.6}& \textbf{10.2} \\
    \bottomrule \\
    \end{tabular}
    \caption{Cross-modal action retrieval results on EPIC including \mm(Caption RNN).}
    \label{tab:cross_modal_results}
\end{table}

\begin{table}[]
\centering
\footnotesize
    \begin{tabular}{lcrrcrr}
    \toprule
    EPIC                    & & \multicolumn{2}{c}{SEEN} & & \multicolumn{2}{c}{UNSEEN}\\
                            & & vv & tt & & vv & tt  \\ \midrule
    Random Baseline         & & 0.6& 0.6& & 0.9&0.9  \\
    CCA Baseline            & &13.8&62.2& &18.9&68.5 \\
    Features(Word2Vec)      & & -- &62.5& & -- &71.3 \\
    Features(Video)         & &13.6& -- & &21.0& --  \\ \midrule
    \mm~(Caption)           & &17.2&63.8& &20.7&69.6 \\
     \mm~(Caption RNN)      & &17.6&73.5& &22.1&76.1 \\
    \mm~([Verb, Noun])      & &17.6&83.5& &22.5&84.7 \\ \midrule
    \jp (Verb,Noun)         & &\textbf{18.8}&\textbf{87.7}& &\textbf{23.2}&\textbf{87.7} \\ 
    \bottomrule \\
    \end{tabular}
    \caption{Within-modal action retrieval results on EPIC including \mm(Caption RNN).}
    \label{tab:within_modal_results}
\end{table}

Cross-Modal and Within-Modal Results can be seen in Tables~\ref{tab:cross_modal_results} and~\ref{tab:within_modal_results} respectively. The inclusion of the RNN sees improvements in mAP performance for tv, vv and tt compared to \mm~(caption). However, compared to \mm~([Verb,Noun]) or \jp~(Verb,Noun) using the entire caption still leads to worse results for both cross and within modal retrieval.

\section{Additional MSR-VTT Experiments \newline (Sec. 4.2)}
Table~\ref{tab:msr_vtt_ext} of this supplementary is an expanded version of Table~\textcolor[rgb]{1,0,0}{7} in the main paper testing a variety of different combinations for PoS.
For each row, an average of 10 runs is reported.
This experiment also includes the removal of the NetVLAD layer in the \mm, substituting it with mean pooling which we label as AVG.

Results show that, on their own, Determinants, Adjectives and Adpositions achieve very poor results.
We also report three \jp{} disentanglement options: (Verb,~Noun), (Caption$\setminus$Verb, Verb) and the one in the main paper (Capiton$\setminus$Noun, Noun).
The table shows that the best results are achieved when nouns are disentangled from the rest of the caption.

\begin{table*}[t]
    \centering
    \begin{tabular}{lcrrrrcrrrr}
    \toprule
                            & & \multicolumn{4}{c}{Video-to-text} & & \multicolumn{4}{c}{Text-to-Video} \\
    MSR-VTT Retrieval       & &  R@1 &  R@5 & R@10 &  MR    & &  R@1 &  R@5 & R@10 &  MR  \\ \midrule
    Mixture of Experts~\cite{miech2018learning}*& &  --    &  --  &  --  &  --  & & 12.9 & 36.4 & 51.8 & 10   \\ \midrule
    Random Baseline          & & 0.3  & 0.7  & 1.1  & 502    & & 0.3  & 0.7  & 1.1  & 502    \\
    CCA Baseline             & & 2.8  & 5.6  & 8.2  & 283    & &  7.0 & 14.4 & 18.7 & 100    \\ \midrule
    \mm(DET AVG)             & & 0.0  & 0.2  & 0.5  & 214    & &  0.3 &  1.0 &  2.2 & 264    \\
    \mm(ADJ AVG)             & & 0.0  & 0.3  & 0.7  & 216    & &  0.1 &  1.1 &  2.6 & 260    \\
    \mm(ADP AVG)             & & 0.1  & 0.6  & 1.5  & 172    & &  0.7 &  2.8 &  5.0 & 185    \\
    \mm(Verb AVG)            & & 1.1  & 5.4  & 11.1 & 57     & &  3.2 & 10.9 & 17.4 &  57    \\
    \mm(Noun AVG)            & & 10.0 & 28.0 & 40.0 & 16     & & 10.7 & 29.7 & 43.5 &  15    \\
    \midrule 
    \mm(DET NetVLAD)         & & 0.0  & 0.1  & 0.3  & 241    & &  0.1 &  1.1 &  2.4 & 255    \\
    \mm(ADJ NetVLAD)         & & 0.0  & 0.0  & 0.1  & 232    & &  0.2 &  1.2 &  2.0 & 262    \\
    \mm(ADP NetVLAD)         & & 0.1  & 0.7  & 1.5  & 174    & &  0.6 &  2.9 &  4.9 & 190    \\
    \mm(Verb NetVLAD)        & & 0.7  & 4.0  & 8.3  &  70    & &  2.9 &  7.9 & 13.9 &  63    \\
    \mm(Noun NetVLAD)        & & 10.8 & 31.3 & 42.7 &  14    & & 10.8 & 30.7 & 44.5 &  13    \\
    \midrule
    \mm([V, N, DET] AVG)     & &  9.0 & 28.4 & 41.0 & 15     & & 7.7  & 24.2 & 36.0 &  20    \\ 
    \mm([Verb,Noun] AVG)     & & 12.9 & 34.0 & 46.7 & 12     & & 12.6 & 32.6 & 46.3 &  12    \\
    \mm([V, N, ADP] AVG)     & & 13.0 & 33.0 & 46.0 & 13     & & 12.2 & 33.0 & 46.0 &  13    \\
    \mm([V, N, ADJ] AVG)     & & 12.4 & 32.9 & 45.3 & 13     & & 11.0 & 31.2 & 44.3 &  13    \\
    \mm([V, N, ADJ, ADP] AVG)& & 13.0 & 32.3 & 45.9 & 12     & & 11.1 & 31.5 & 44.3 &  13    \\
    \midrule
    \mm([V, N, DET] NetVLAD) & & 14.8 & 38.3 & 52.5 & 9.1    & & 12.4 & 33.6 & 46.3 &  13    \\
    \mm([Verb,Noun] NetVLAD) & & 15.6 & 39.4 & 55.1 & 9.0    & & 13.6 & 36.8 & 51.7 &  10    \\
    \mm([V, N, ADP] NetVLAD) & & 15.8 & 40.3 & 55.1 &8.5& & 13.8 & 36.7 & 51.0 &  10    \\
    \mm([V, N, ADJ] NetVLAD) & & 16.3 & 40.1 & 54.1 & 8.9    & & 14.0 & 36.2 & 50.9 &  10    \\ 
    \mm([V, N, ADJ, ADP] NetVLAD) & & 16.1 & 39.7 & 53.8 & 8.9& &13.4  & 36.2 & 51.3 & 10    \\ 
    \midrule
    \mm(Caption AVG)         & & 12.4 & 32.8 & 45.6 &  12    & & 11.4 & 31.2 & 43.8 &  14    \\
    \mm(Caption NetVLAD)     & & 15.8 & 40.2 & 53.6 &   9    & & 13.8 & 36.7 & 50.7 &  10.3  \\
    \midrule
    \jp(Verb, Noun) & &  15.5&39.3  & 53.8 & 9      & & 13.7 & 37.6 & 52.2 & 9.6    \\
    \jp(Caption$\setminus$Verb,Verb)& & 15.9 & 39.2 &{\bf55.5}&{\bf8}  & & 13.4 & 36.8 & 52.0 &  10    \\
    \jp(Caption$\setminus$Noun,Noun)& &{\bf16.4}&{\bf41.3}& 54.4 &   8.7  & &{\bf14.3}&{\bf38.1}&{\bf53.0}&{\bf9}  \\
    \bottomrule \\
    \end{tabular}
    \caption{MSR-VTT Video-Caption Retrieval results using recall@k (R@k, higher is better) and median Rank (MR, lower is better). For each row, an average of 10 runs is reported. *We include results from ~\cite{miech2018learning}, only available for Text-to-Video retrieval.} 
    \label{tab:msr_vtt_ext}
\end{table*}

\end{document}